\documentclass[runningheads]{llncs}
\usepackage[nolist]{acronym}
\usepackage{graphics}
\usepackage{graphicx}
\usepackage{url}
\usepackage{longtable}
\usepackage{supertabular}
\usepackage{pdflscape}
\usepackage{pifont}

\usepackage{rotating}
\usepackage{array}
\usepackage{tabularx}
\usepackage{lscape}
\usepackage{multirow}
\usepackage{multicol}
\usepackage{pdflscape}
\usepackage{mathtools}
\usepackage{footnote}
\usepackage{verbatim}
\usepackage{supertabular}
\usepackage{hyperref}
\begin{acronym}[UML]
	\acro{AOS}{Agricultural Ontology Services}
	\acro{AGRIS}{Agricultural Science and Technology}
	\acro{API}{Application Programming Interface}
	\acro{BPSO}{Binary Particle-Swarm Optimization}
	\acro{BPMLOD}{Best Practices for Multilingual Linked Open Data}
	\acro{CBD}{Concise Bounded Description}
	\acro{COG}{Content Oriented Guidelines}
	\acro{CSV}{Comma-Separated Values}
	\acro{CBMT}{Corpus-Based Machine Translation}
	\acro{CLIR}{Cross-Language Information Retrieval}
	\acro{DPSO}{Deterministic Particle-Swarm Optimization}
	\acro{DALY}{Disability Adjusted Life Year}

	\acro{ER}{Entity Resolution}
	\acro{EM}{Expectation Maximization}
	\acro{EBMT}{Example-Based Machine Translation}
	\acro{EBNF}{Extended Backus--Naur Form}
	\acro{EL}{Entity Linking}
	\acro{FAO}{Food and Agriculture Organization of the United Nations}
	\acro{GIS}{Geographic Information Systems}
	\acro{GHO}{Global Health Observatory}
	\acro{HDI}{Human Development Index}
	\acro{ICT}{Information and communication technologies}
    \acro{KB}{Knowledge Base}
    \acro{KG}{Knowledge Graph}
    \acro{KBSE}{Knowledge Base Semantic Embedding}
	\acro{LR}  {Language Resource}
	\acro{LD}  {Linked Data}
	\acro{LLOD}  {Linguistic Linked Open Data}
	\acro{LIMES}{LInk discovery framework for MEtric Spaces}
	\acro{LS}  {Link Specifications}
	\acro{LDIF}{Linked Data Integration Framework}
	\acro{LGD} {LinkedGeoData}
	\acro{LOD} {Linked Open Data}
	\acro{MSE}{Mean Squared Error}
	\acro{MWE}{Multiword Expressions}
	\acro{MT}{Machine Translation}
	\acro{ML}{Machine Learning}
	\acro{NIF}{Natural Language Processing Interchange Format}
	\acro{NIF4OGGD}{NLP Interchange Format for Open German Governmental Data}
	\acro{NLP}{Natural Language Processing}
	\acro{NER}{Named Entity Recognition}
	\acro{NMT}{Neural Machine Translation}
	\acro{NN}{Neural Network}
	\acro{NLG}{Natural Language Generation}
	\acro{NED}{Named Entity Disambiguation}
	\acro{NERD}{Named Entity Recognition and Disambiguation}
	\acro{NL}{Natural Language}
	\acro{OSM}{OpenStreetMap}
	\acro{OWL}{Web Ontology Language}
	\acro{OOV}{out-of-vocabulary}
	\acro{PFM}{Pseudo-F-Measures}
	\acro{PSO}{Particle-Swarm Optimization}
	\acro{QA}{Question Answering}
	\acro{RDF}{Resource Description Framework}
	\acro{RBMT}{Ru\-le-Ba\-sed Ma\-chi\-ne Trans\-la\-tion}
	\acro{SKOS}{Simple Knowledge Organization System}
	\acro{SPARQL}{SPARQL Protocol and RDF Query Language}
	\acro{SRL}{Statistical Relational Learning}
	\acro{SWT}{Semantic Web Technologies}
	\acro{SW}{Semantic Web}
	\acro{SMT}{Statistical Machine Translation}
	\acro{SWMT}{Semantic Web Machine Translation}

    \acro{TBMT} {Transfer-Based Machine Translation}
	\acro{UML}{Unified Modeling Language}
	\acro{USL}{Ukrainian Sign Language}
	\acro{WHO}{World Health Organization}
	\acro{WKT}{Well-Known Text}
	\acro{W3C}{World Wide Web Consortium}
	\acro{WSD}{Word Sense Disambiguation}
    \acro{XML}{Extensible Markup Language}
	\acro{YPLL}{Years of Potential Life Lost}
\end{acronym}  
\begin{document}
\title{Semantic Web for Machine Translation: Challenges and Directions}

\author{Diego Moussallem\inst{1}\and
Matthias Wauer\inst{1}\and
Axel-Cyrille Ngonga Ngomo\inst{1}}
\authorrunning{Moussallem et al.}
%
\institute{Data Science Group, University of Paderborn, Germany \\
\email{first.lastname@upb.de}\\ 
}
\maketitle              
\begin{abstract}

A large number of machine translation approaches have recently been developed to facilitate the fluid migration of content across languages. However, the literature suggests that many obstacles must still be dealt with to achieve better automatic translations. One of these obstacles is lexical and syntactic ambiguity. A promising way of overcoming this problem is using Semantic Web technologies. This article is an extended abstract of our systematic review on machine translation approaches that rely on Semantic Web technologies for improving the translation of texts. Overall, we present the challenges and opportunities in the use of Semantic Web technologies in Machine Translation. Moreover, our research suggests that while Semantic Web technologies can enhance the quality of machine translation outputs for various problems, the combination of both is still in its infancy.

\end{abstract}

\section{Introduction}

Alongside increasing globalization comes a greater need for readers to understand texts in languages foreign to them. For example, approximately 48\% of the pages on the Web are not available in English\footnote{\url{https://www.internetworldstats.com/stats7.htm}}. The technological progress of recent decades has made both the distribution and access to content in different languages ever simpler. Translation aims to support users who need to access content in a language in which they are not fluent~\cite{Koehn2010}.

However, translation is a difficult task due to the complexity of natural languages and their structure~\cite{Koehn2010}. In addition, manual translation does not scale to the magnitude of the Web. One remedy for this problem is \ac{MT}. The main goal of \ac{MT} is to enable people to assess content in languages other than the languages in which they are fluent~\cite{Bar-Hillel1960}. From a formal point of view, this means that the goal of \ac{MT} is to transfer the semantics of text from an input language to an output language~\cite{hutchins1992introduction}.

Although \ac{MT} systems are now popular on the Web, they still generate a large number of incorrect translations. Recently, Popovi{\'c}~\cite{popovic2012class} has classified five types of errors that still remain in \ac{MT} systems. According to research, the two main faults that are responsible for 40\% and 30\% of problems respectively, are reordering errors and lexical and syntactic ambiguity. Thus, addressing these barriers is a key challenge for modern translation systems. A large number of \ac{MT} approaches have been developed over the years that could potentially serve as a remedy. For instance, translators began by using methodologies based on linguistics which led to the family of \ac{RBMT}. However, \ac{RBMT} systems have a critical drawback in their reliance on manually crafted rules, thus making the development of new translation modules for different languages even more difficult.

\ac{SMT} and \ac{EBMT} were developed to deal with the scalability issue in \ac{RBMT}~\cite{brown1990statistical}, a necessary characteristic of \ac{MT} systems that must deal with data at Web scale. Presently, these approaches have begun to address the drawbacks of rule-ba\-sed approaches. However, some problems that had already been solved for linguistics based methods reappeared. The majority of these problems are connected to the issue of ambiguity, including syntactic and semantic variations~\cite{Koehn2010}. Nowadays, a novel \ac{SMT} paradigm has ari\-sen called \ac{NMT} which relies on \ac{NN} algorithms. \ac{NMT} has been achieving impressive results and is now the state-of-the-art in \ac{MT} approaches. However, \ac{NMT} is still a statistical approach sharing some semantic drawbacks from other well-defined \ac{SMT} approaches\cite{koehn2017six}.  

One possible solution to address the remaining issues of \ac{MT} lies in the use of \ac{SWT}, which have emerged over recent decades as a paradigm to make the semantics of content explicit so that it can be used by machines. It is believed that explicit semantic knowledge made available through these technologies can empower \ac{MT} systems to supply translations with significantly better quality while remaining scalable. In particular, the disambiguated knowledge about real-world entities, their properties and their relationships made available on the \ac{LD} Web can potentially be used to infer the right meaning of ambiguous sentences or words.

According to our survey~\cite{moussallem2018machine}, the obvious opportunity of using \ac{SWT} for \ac{MT} has already been studied by a number of approaches, especially w.r.t. the issue of ambiguity. In this paper, we present the challenges and opportunities in the use of \ac{SWT} in \ac{MT} for translating texts.

\section{Related Works}

The idea of using a structured \ac{KB} in \ac{MT} systems started in the 90s with the work of Knight and Luk~\cite{knight1994building}. Still, only a few researchers have designed different strategies for benefiting of structured knowledge in \ac{MT} architectures~\cite{arcan2015knowledge}. Recently, the idea of using \ac{KG} into \ac{MT} systems has gained renewed attention. Du et al.~\cite{jinhua2016} created an approach to address the problem of \ac{OOV} words by using BabelNet~\cite{navigli2012babelnet}. Their approach applies different methods of using BabelNet. In summary, they create additional training data and apply a post-editing technique, which replaces the \ac{OOV} words while querying BabelNet. Shi et al.~\cite{shi2016knowledge} have recently built a semantic embedding model reliant upon a specific \ac{KB} to be used in \ac{NMT} systems. The model relies on semantic embeddings to encode the key information contained in words to translate the meaning of sentences correctly. The work consists of mapping a source sentence to triples, which are then used to extract the intrinsic meaning of words to generate a target sentence. This mapping results in a semantic embedding model containing \ac{KB} triples, which are responsible for gathering the key information of each word in the sentences.  

\section{Open MT Challenges}
\label{sec:openMTproblems}

The most problematic unresolved \ac{MT} challenges, from our point of view, which are still experienced by the aforementioned \ac{MT} approaches are the following:

\begin{enumerate}

\item  \emph{Complex semantic ambiguity}: This challenge is mostly caused by the existence of ho\-mo\-ny\-mous and po\-ly\-se\-mous words. Given that a significant amount of parallel data is necessary to translate such words and expressions adequately. \ac{MT} systems commonly struggle to translate these words correctly, even if the models are built upon from 5- or 7-grams. For exam\-ple, ``John \underline{promises to keep} his room tidy" and ``John has some \underline{promises to keep} until he is trusted again". Although the meaning of both are clear to humans, these sentences for \ac{SMT} systems are statistically expensive and prone to failure\footnote{See a complete discussion about the problem:~\url{http://tinyurl.com/yck5ngj8}}. Additionally, for translating the simple word ``bank", context information is essential for determining which meaning to assign to it. 

\item  \emph{Structural divergence}: By definition, structural reordering is reorganizing the order of the syntactic constituents of a language according to its original structure.
It in turn becomes a critical issue because it is the core of the translation process. Every language has its own syntax, thus each \ac{MT} system needs to have adequate models for the syntax of each language. For instance, reordering a sentence from Japanese to English is one of the most challenging techniques because of the SVO (sub\-ject-verb-ob\-je\-ct) and SOV (sub\-ject-object-verb) word-order  difference and also, one English word often groups multiple meanings of Japanese characters. For example, Japanese characters make subtle distinctions between homonyms that would not be clear in a phonetic language such as English.

\item  \emph{Linguistic properties/features}: A large number of languages display a complex tense system. When con\-fron\-ted with sentences from such languages, it can be hard for \ac{MT} systems to recognize the current input tense and to translate the input sentence into the right tense in the target language. For instance, some irregular verbs in English like ``set'' and ``put'' cannot be determined to be in the present or past tense without previous knowledge or pre-processing techniques when translated to morphologically rich languages, e.g., Portuguese, German or Slavic languages. Additionally, the grammatical gender of words in such morphologically rich languages contributes to the problem of tense generation where a certain \ac{MT} system has to decide which inflection to use for a given word. This challenge is a direct consequence of the structural reordering issue and remains a significant problem for modern translator systems.
\end{enumerate}

Additionally, there are five \ac{MT} open challenges posed by Lopez and Post~\cite{lopez2013beyond} which we describe more generically below. 

(1) Excessive focus on English and European languages as one of the involved languages in \ac{MT} approaches and poor research on low-resource language pairs such as African and/or South American languages. (2) The limitations of \ac{SMT} approa\-ches for translating across domains. Most \ac{MT} systems exhibit good performance on law and the legislative domains due to the large amount of data provided by the European Union. In contrast, translations performed on sports and life-hacks commonly fail, because of the lack of training data. (3) How to translate the huge amount of data from social networks that uniquely deal with no-standard speech texts from users (e.g., tweets). (4) The difficult translations among morphologically rich languages. This challenge shares the same problem with the first one, namely that most research work focuses on English as one of the involved languages. Therefore, \ac{MT} systems which translate content between, for instance, Arabic and Spanish are rare. (5) For the speech translation task, the parallel data for training differs widely from real user speech.

The challenges above are clearly not independent, which means that addressing one of them can have an impact on the others. Since \ac{NMT} has shown impressive results on reordering, the main problem turns out to be the disambiguation process (both syntactically and semantically) in \ac{SMT} approaches~\cite{Koehn2010}.

\section{Suggestions and Possible Directions using SW}

Based on the surveyed works on our research~\cite{moussallem2018machine}, \ac{SWT} have mostly been applied at the semantic analysis step, rather than at the other stages of the translation process, due to their ability to deal with concepts behind the words and provide knowledge about them. As \ac{SWT} have developed, they have increasingly been able to resolve some of the open challenges of \ac{MT}. They may be applied in different ways according to each \ac{MT} approach. 

\textit{\textbf{Disambiguation.}} Human language is very ambiguous. Most words have multiple interpretations depending on the context in which they are mentioned. In the \ac{MT} field, \ac{WSD} techniques are concerned with finding the respective meaning and correct translation to these ambiguous words in target languages. This ambiguity problem was identified early in \ac{MT} development. In 1960 Bar-Hillel~\cite{Bar-Hillel1960} stated that an \ac{MT} system is not able to find the right meaning without a specific knowledge. Although the ambiguity problem has been lessened significantly since the contribution of Carpuat and subsequent works~\cite{carpuat2007improving}, this problem still remains a challenge.
As seen in Moussallem et al.~\cite{moussallem2018machine}, \ac{MT} systems still try to resolve this problem by using domain specific language models to prefer domain specific expressions, but when translating a highly ambiguous sentence or a short text which covers multiple domains, the languages models are not enough. 

\ac{SW} has already shown its capability for semantic disambiguation of po\-ly\-se\-mous and ho\-mo\-ny\-mous words. However, \ac{SWT} were applied in two ways to support the semantic disambiguation in \ac{MT}. First, the ambiguous words were recognized in the source text before carrying out the translation, applying a pre-editing technique. Second, \ac{SWT} were applied to the output translation in the target language as a post-editing technique. Although applying one of these techniques has increased the quality of a translation, both techniques are tedious to implement when they have to translate common words instead of named entities, then be applied several times to achieve a successful translation.

The real benefit of \ac{SW} comes from its capacity to provide unseen knowledge about emergent data, which appears every day. Therefore, we suggest performing the topic-modelling technique over the source text to provide a necessary context before translation. Instead of applying the topic-modeling over the entire text, we would follow the principle of communication (i.e from 3 to 5 sentences for describing an idea and define a context for each piece of text. Thus, at the execution of a translation model in a given \ac{SMT}, we would focus on every word which may be a homonymous or polysemous word. For every word which has more than one translation, a SPARQL query would be required to find the best combination in the current context. Thus, at the translation phase, the disambiguation algorithm could search for an appropriate word using different \ac{SW} resources such as DBpedia, in consideration of the context provided by the topic modelling. The goal is to exploit the use of more than one \ac{SW} resource at once for improving the translation of ambiguous terms. The use of two or more \ac{SW} resources simultaneously has not yet been investigated.    

On the other hand, there is also a syntactic disambiguation problem which as yet lacks good solutions. For instance, the English language contains irregular verbs like ``set'' or ``put''. Depending on the structure of a sentence, it is not possible to recognize their verbal tense, e.g., present or past tense. Even statistical approaches trained on huge corpora may fail to find the exact meaning of some words due to the structure of the language. Although this challenge has successfully been dealt with since \ac{NMT} has been used for European languages, implementations of \ac{NMT} for some non-European languages have not been fully exploited (e.g., Brazilian Portuguese, Latin-America Spanish, Zulu, Hindi) due to the lack of large bilingual data sets on the Web to be trained on. Thus, we suggest gathering relationships among properties within an ontology by using the reasoning technique for handling this issue. For instance, the sentence ``Anna usually put her notebook on the table for studying" may be annotated using a certain vocabulary and represented by triples. Thus, the verb ``put", which is represented by a predicate that groups essential information about the verbal tense, may support the generation step of a given \ac{MT} system. This sentence usually fails when translated to rich morphological languages, such as Brazilian-Portuguese and Arabic, for which the verb influences the translation of ``usually" to the past tense. In this case, a reasoning technique may support the problem of finding a certain rule behind relationships between source and target texts in the alignment phase (training phase). However, a well-known problem of reasoners is the poor run-time performance. Therefore, this run-time deficiency needs to be addressed or minimized before implementing reasoners successfully into \ac{MT} systems.     

\textit{\textbf{Named Entities.}} Most \ac{NERD} approaches link recognized entities with database entries or websites. This method helps to categorize and summarize text, but also contributes to the disambiguation of words in texts. The primary issue in \ac{MT} systems is caused by common words from a source language that are used as proper nouns in a target language. For instance, the word ``Kiwi" is a family name in New Zealand which comes from the M\=aori culture, but it also can be a fruit, a bird, or a computer program. Named Entities are a common and difficult problem in both \ac{MT} (see Koehn~\cite{Koehn2010}) and \ac{SW} fields. The \ac{SW} achieved important advances in \ac{NERD} using structured data and semantic annotations, e.g., by adding an \texttt{rdf:type} statement which identifies whether a certain kiwi is a fruit~\cite{moussallem2017mag}. In \ac{MT} systems, however, this problem is directly related to the ambiguity problem and therefore has to be resolved in that wider context.

Although \ac{MT} systems include good recognition methods, they still need improvement. When an \ac{MT} system does not recognize an entity, the translation output often has poor quality, immediately deteriorating the target text readability. 
Therefore, we suggest recognizing such entities before the translation process and first linking them to a reference knowledge base. Afterwards, the type of entities would be agglutinated along with their labels and their translations from a reference knowledge base. For instance, in \ac{NMT}, the idea is to include in the training set for the aforementioned word ``Kiwi", ``Ki\-wi.ani\-mal.link, Ki\-wi.per\-son.link, Ki\-wi.fo\-od.link" then finally to align them with the translations in the target text. For example, in \ac{SMT}, the additional information can be included by \ac{XML} or by an additional model. In contrast, in \ac{NMT}, this additional information can be used as parameters in the training phase. This method would also contribute to \ac{OOV} mistakes regarding names. This idea is supported by \cite{shi2016knowledge} where the authors encoded the types of entities along with the words to improve the translation of sentences between Chinese-English. Recently, Moussallem et al.~\cite{moussallem2019augmenting} have shown promising results by applying a multilingual entity linking algorithm along with knowledge graph embeddings into the translation phase of a neural machine translation model for improving the translation of entities in texts. Their approach achieved significant and consistent improvements of +3 BLEU, METEOR and CHRF3 on average on the newstest datasets between 2014 and 2018 for WMT English-German translation task. 

\textit{\textbf{Non-standard speech.}} The non-standard language problem is a rather important one in the \ac{MT} field. Many people use the colloquial form to speak and write to each other on social networks. Thus, when \ac{MT} systems are applied on this context, the input text frequently contains slang, \ac{MWE}, and unreasonable abbreviations such as
``Idr = I don't remember.'' and ``cya = see you''. Additionally, idioms contribute to this problem, decreasing the translation quality. Idioms often have an entirely different meaning than their separated word meanings. Consequently, most translation outputs of such expressions contain errors. For a good translation, the \ac{MT} system needs to recognize such slang and try to map it to the target language. Some \ac{SMT} systems like Google or Bing have recognition patterns over non-standard speech from old translations through the Web using \ac{SMT} approaches. In rare cases \ac{SMT} can solve this problem, but considering that new idiomatic expressions appear every day and most of them are isolated sentences, this challenge still remains open. Moreover, each person has their own speaking form.

Therefore, we suggest that user characteristics can be applied as context for solving the non-standard language problem. These characteristics can be extracted from social media or user logs and stored as user properties using \ac{SWT}, e.g., FOAF vocabulary. These ontologies have properties which would help identify the birth place or the interests of a given user. For instance, the properties \emph{foaf:interest} and \emph{sioc:topic} can be used to describe a given person's topics of interest. If the person is a computer scientist and the model contains topics such as ``Information Technology" and ``Sports", the SPARQL queries would search for terms inserted in this context which are ambiguous. Furthermore, the property \emph{foaf:based\_near} may support the problem of idioms. Assuming that a user is located in a certain part of Russia and he is reading an English web page which contains some idioms, this property may be used to gather appropriate translations of idioms from English to Russian using a given RDF \ac{KB}. Therefore, an \ac{MT} system can be adapted to a user by using specific data about him in \ac{RDF} along with given \ac{KB}s. Recently, Moussallem et al~\cite{moussallemlrec2018} have released a multilingual linked idioms dataset as a first part of supporting the investigation of this suggestion. The dataset contains idioms in 5 languages and are represented by knowledge graphs which facilitates the retrieval and inference of translations among the idioms.

\textit{\textbf{Translating \ac{KB}s.}} According to our research, it is clear that \ac{SWT} may be used for translating \ac{KB}s in order to be applied in \ac{MT} systems. For instance, some content provided by the German Wikipedia version are not contained in the Portuguese one. Therefore, the semantic structure (i.e., triples) provided by DBpedia versions of these respective Wikipedia versions would be able to help translate from German to Portuguese. For example, the terms contained in triples would be translated to a given target language using a dictionary containing domain words. This dictionary may be acquired in two different ways. First, by performing localisation, as in the work by J. P. McCrae~\cite{mccrae2016domain} which translates the terms contained in a monolingual ontology, thus generating a bilingual ontology. Second, by creating embeddings of both DBpedia versions in order to determine the similarity between entities through their vectors. This insight is supported by some recent works, such as Ristoski et al.~\cite{ristoski2016rdf2vec}, which creates bilingual embeddings using \ac{RDF} based on Word2vec algorithms. Therefore, we suggest investigating an \ac{MT} approach mainly based on \ac{SWT} using \ac{NN} for translating \ac{KB}s. Once the \ac{KB}s are translated, we suggest including them in the language models for improving the translation of entities. 

Besides C. Shi et al~\cite{shi2016knowledge}, Ar\v{c}an and Buitelaar~\cite{arcan2017translating} presented an approach to translate domain-specific expressions represented by English \ac{KB}s in order to make the knowledge accessible for other languages. They claimed that \ac{KB}s are mostly in English, therefore they cannot contribute to the problem of \ac{MT} to other languages. Thus, they translated two \ac{KB}s belonging to medical and financial domains, along with the English Wikipedia, to German. Once translated, the \ac{KB}s were used as external resources in the translation of German-English. The results were quite appealing and the further research into this area should be undertaken. Recently, Moussallem et al~ \cite{moussallem2019thoth} created THOTH, an approach which translates and enriches knowledge graphs across languages. Their approach relies on two different recurrent neural network models along with knowledge graph embeddings. The authors applied their approach on the German DBpedia with the German translation of the English DBpedia on two tasks: fact checking and entity linking. THOTH showed promising results with a translation accuracy of 88.56 while being capable of improving two NLP tasks with its enriched-German \ac{KG} .

\vspace{-2mm}
\section{conclusion}

In this extended abstract, we detailed the results of a systematic literature review of \ac{MT} using \ac{SWT} for improving the translation of natural language sentences. Our goal was to present the current open \ac{MT} translation problems and how \ac{SWT} can address these problems and enhance \ac{MT} quality. Considering the decision power of \ac{SWT}, they cannot be ignored by future \ac{MT} systems. As a next step, we intend to continue elaborating a novel \ac{MT} approach which is capable of simultaneously gathering knowledge from different \ac{SW} resources and consequently being able to address the ambiguity of named entities and also contribute to the \ac{OOV} words problem. This insight relies on our recent works, such as \cite{moussallem2019augmenting}, which have augmented \ac{NMT} models with the usage of external knowledge for improving the translation of entities in texts. Additionally, future works that can be expected from fellow researchers, include the creation of multilingual linguistic ontologies describing the syntax of rich morphologically languages for supporting \ac{MT} approaches. Also, the creation of more \ac{RDF} multilingual dictionaries which can improve some \ac{MT} steps, such as alignment.

\section*{Acknowledgments} 	

This work was supported by the German Federal Ministry of Transport and Digital Infrastructure (BMVI) in the projects LIMBO (no. 19F2029I) and OPAL (no. 19F2028A) as well as by the Brazilian National Council for Scientific and Technological Development (CNPq) (no. 206971/2014-1)

\bibliographystyle{splncs04}
\bibliography{mybibliography}

\end{document}